\documentclass[pmlr]{jmlr}

\usepackage{booktabs}
\usepackage[load-configurations=version-1]{siunitx} 

\theorembodyfont{\upshape}
\theoremheaderfont{\scshape}
\theorempostheader{:}
\theoremsep{\newline}

\jmlrvolume{XX}
\jmlryear{2019}
\jmlrworkshop{Machine Learning for Health (ML4H) at NeurIPS 2019}

\usepackage[utf8]{inputenc} 
\usepackage[T1]{fontenc}    
\usepackage{url}            
\usepackage{booktabs}       
\usepackage{amsfonts}       
\usepackage{nicefrac}       
\usepackage{microtype}      
\usepackage{amsmath}
\usepackage{algorithm}
\usepackage{algorithmic}

\usepackage{color}
\usepackage{float}
\usepackage{enumitem}
\usepackage{amsfonts}
\usepackage{amssymb}
\usepackage{url}
\usepackage{amsfonts}
\usepackage{color}
\usepackage{multicol}
\usepackage{comment}
\usepackage{lipsum}
\usepackage{afterpage}
\usepackage{amsmath}
\usepackage{mathrsfs}

\usepackage{graphicx}

\title[DermGAN]{DermGAN:\\ Synthetic Generation of Clinical Skin Images with Pathology}

 \author{\Name{Amirata Ghorbani}\thanks{Work done while interning at Google Health.} \Email{amiratag@stanford.edu}\\
 \addr Department of Electrical Engineering \\
       Stanford, CA
 \AND
 \Name{Vivek Natarajan} \Email{natviv@google.com}\\
 \Name{David Coz} \Email{dcoz@google.com}\\
 \Name{Yuan Liu} \Email{yuanliu@google.com}\\
 \addr Google Health \\
 Palo Alto, CA
 }
 
\begin{document}

\maketitle

\begin{abstract}

    Despite the recent success in applying supervised deep learning to medical imaging tasks, the problem of obtaining large and diverse expert-annotated datasets required for the development of high performant models remains particularly challenging. In this work, we explore the possibility of using Generative Adverserial Networks (GAN) to synthesize clinical images with skin condition. We propose DermGAN, an adaptation of the popular Pix2Pix architecture, to create synthetic images for a pre-specified skin condition while being able to vary its size, location and the underlying skin color. We demonstrate that the generated images are of high fidelity using objective GAN evaluation metrics. In a Human Turing test, we note that the synthetic images are not only visually similar to real images, but also embody the respective skin condition in dermatologists' eyes. Finally, when using the synthetic images as a data augmentation technique for training a skin condition classifier, we observe that the model performs comparably to the baseline model overall while improving on rare but malignant conditions.
    \end{abstract}

\section{Introduction}

     The combination of large scale data and compute has catalyzed the success of supervised deep learning in many domains including computer vision \citep{mahajan2018exploring}, natural language processing \citep{devlin2018bert} and speech recognition \citep{hannun2014deep}. Over the last few years, several efforts have been made to apply supervised deep learning to various medical imaging tasks such as disease classification, detection of suspicious malignancy and organ segmentation on different imaging modalities including ophthalmology, pathology, radiology, cardiology, and dermatology~\citep{esteva2017dermatologist,ghorbani2019deep,gulshan2016development,ardila2019end,rajpurkar2017chexnet}. Despite this progress, developing effective deep learning models for these tasks remain non trivial mainly due to the data hungry nature of such algorithms. Most previous efforts that report expert-level performance required large amounts of expert annotated data (multiple thousands and sometimes even millions of training examples). However, the cost of obtaining expert-level annotations in medical imaging is often prohibitive. Moreover, it is near impossible to collect diverse datasets that are unbiased and balanced. Most of the data used in medical imaging and other healthcare applications come from medical sites which may disproportionately serve certain specific demographics. Such datasets also tend to have very few examples of rare conditions because they naturally occur sparingly in the real world. Models trained on such biased and unbalanced datasets tend to perform poorly on test cases drawn from under-represented populations or on rare conditions~\citep{adamson2018machine}. To ameliorate these issues, generative models present an intriguing alternative to fill this data void.
    
    There has been remarkable progress in generative models in recent years. Generative Adverserial Networks (GAN) \citep{goodfellow2014generative}, in particular, have emerged as the de facto standard for generating diverse and high quality samples, with many popular extensions such as CycleGAN \citep{zhu2017unpaired}, StarGAN \citep{choi2018stargan}, StyleGAN \citep{karras2019style} and BigGAN \citep{brock2018large}. GAN have been effectively used in many applications, including super resolution \citep{ledig2017photo}, text-to-image generation \citep{zhang2017stackgan} and image in-painting \citep{pathakCVPR16context}. In the medical domain, applications include generating medical records \citep{choi2017generating}, liver lesion images \citep{frid2018gan}, bone lesions \citep{gupta2019generative} and anomaly detection \citep{zenati2018efficient}.
    
    In this work, we explore the possibility of synthesizing images of skin conditions that were taken by consumer grade cameras. We formulate the problem as an image to image translation task and use an adapted version of the existing GAN-based image translation architectures~\citep{pix2pix, pix2pixhd}. Specifically, our model learns to translate a semantic map with a pre-specified skin condition, its size and location, and the underlying skin color, to a realistic image that preserves the pre-specified traits. In this way, images of rare skin conditions in minority demographics can be generated to diversify existing datasets for the downstream skin condition classification task. We demonstrate via both GAN evaluation metrics and qualitative tests that the generated images are of high fidelity and represent the respective skin condition. When we use the synthetic images as additional data to train a skin condition classifier, we observe that the model improves on rare malignant classes while being comparable to the baseline model overall.

\paragraph{Motivation and distinction from related work}

\begin{figure}[h]
  \centering
  \includegraphics[width=\linewidth]{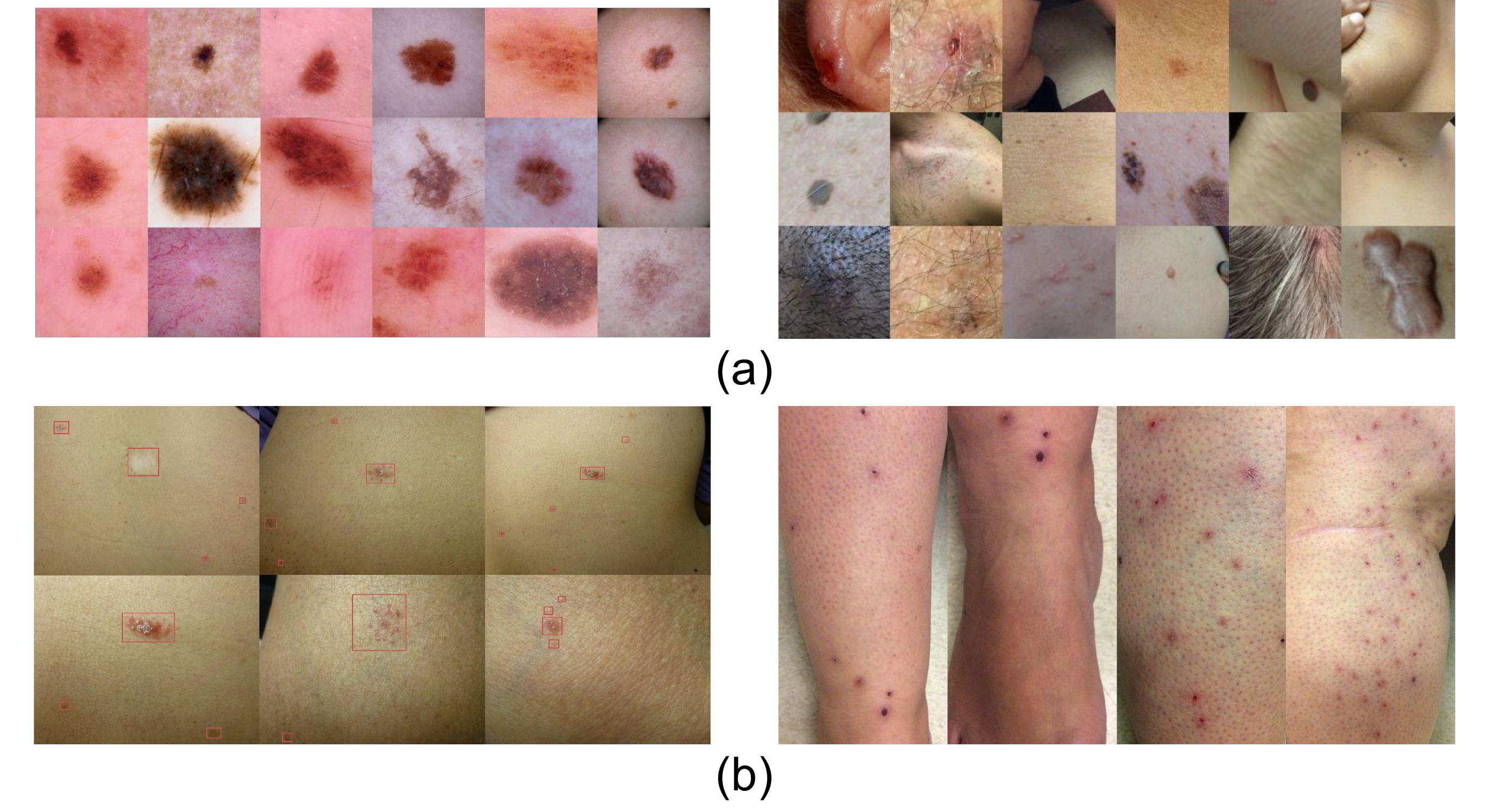}
  \caption{\textbf{Sample skin images in different datasets}  Comparison between samples of dermatoscopic images (left)~\citep{ham1000} and samples of cropped clinical images in our dataset (right) 
    are shown in (a). Our original, uncropped images are shown in (b), collected from a real-world teledermatology service with varying size, scale, and quality.}
   \label{fig:comparison}
\end{figure}

In dermatology, prior efforts~\citep{prev1, prev2, prev3} on applying generative models to synthesize images have focused on datasets of dermoscopic images~\citep{isic, ham1000}. Dermoscopic images are acquired using specialized equipment (dermatoscopes) in order to have a clean, centered, and zoomed-in image of the skin condition under normalized lighting. However, access to dermatoscopes is limited: they are often only available in dermatology clinics and are used to examine certain lesion conditions. On the other hand, clinical images are taken by consumer grade cameras (point-and-shoot cameras or smartphones), and are thus much more accessible to general users. Such images can be used either in a teledermatology setting, where patients or general practitioners can send such photographs to remote dermatologists for diagnosis, or to directly leverage AI based tools for self diagnosis. However, acquisition of such images is not part of the standard clinical workflow, leading to a data void to develop performant skin disease classification models \citep{yang2018clinical, mishra2019interpreting}. Last but not least, unlike dermoscopy images, clinical images of skin conditions have diverse appearances in terms of scale, perspective, zoom effects, lighting, blur and other imaging artifacts. In addition, presence of hair, various skin colors, and body parts, age induced artifacts (e.g., wrinkles), and background also contribute to the diversity of clinical data. Such diversity makes it challenging for generative models to learn the underlying image representation. Fig.~\ref{fig:comparison}(a) contrasts examples of a dermascopy dataset (on the left) ~\citep{ham1000} to that of a clinical dataset (on the right). To the best of our knowledge, no prior work has attempted to synthesize clinical images with skin pathology.

\section{Methods}

\begin{figure}[h]
  \centering
  \includegraphics[width=0.75\linewidth]{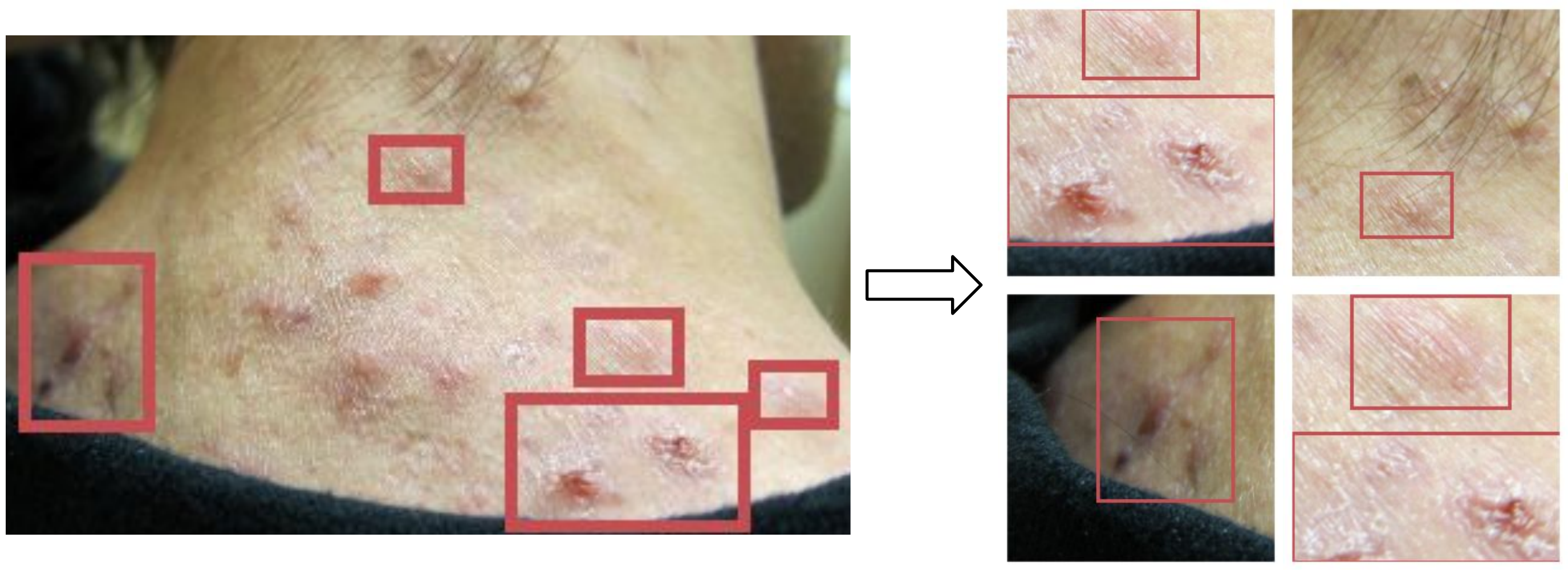}
  \caption{\textbf{Image pre-processing} Using the human-provided ROI annotation of skin conditions, we are able to create cropped images with clear skin condition in focus.
  \label{fig:cropping}} 
\end{figure}

\begin{figure}[h]
  \centering
  \includegraphics[width=\linewidth]{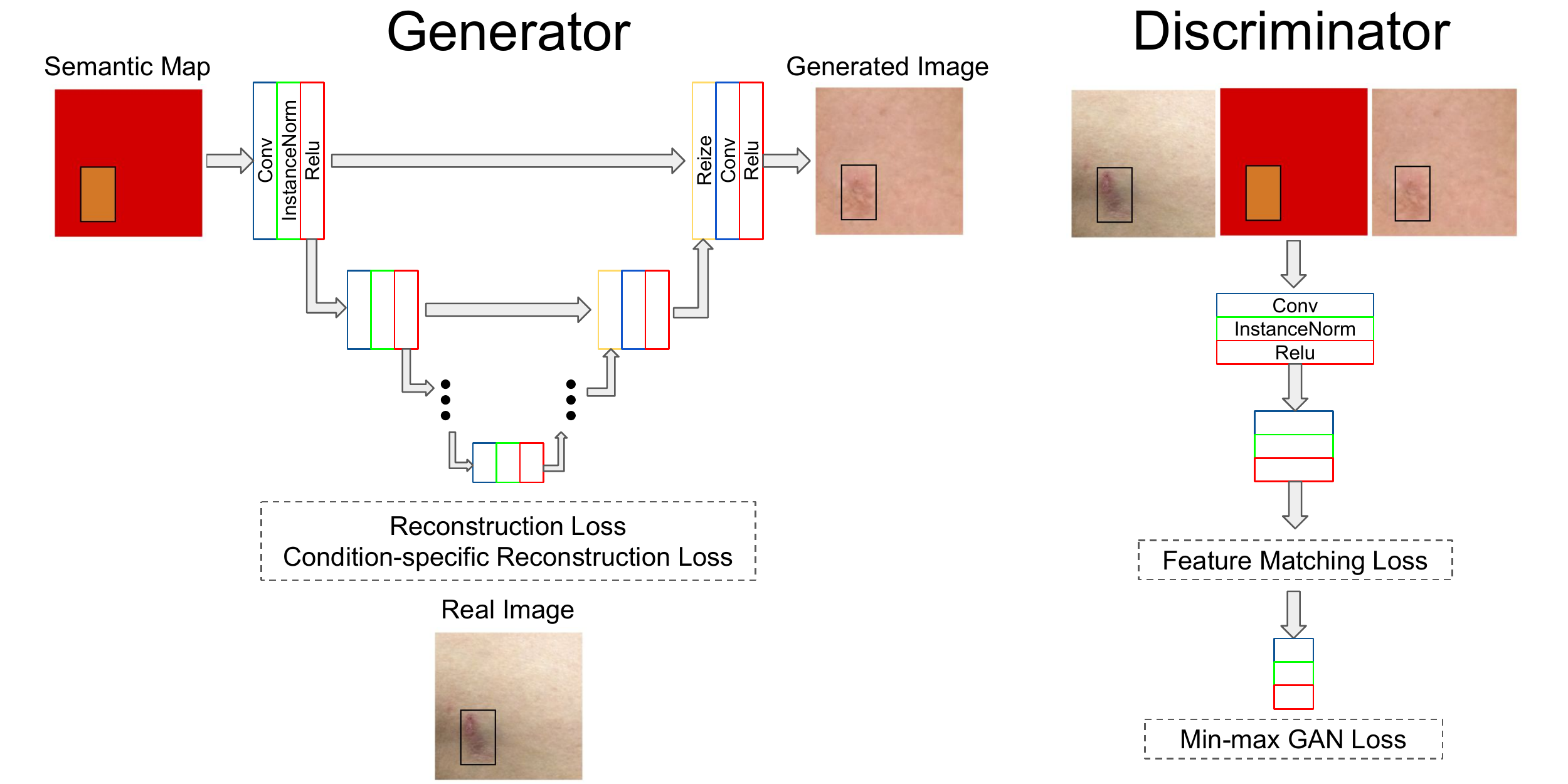}
  \caption{\textbf{DermGAN architecture} A semantic map encoding the skin condition and its region of presence (orange rectangle) and the skin color (red background) is passed through the generator to produce a synthetic image. The generator is a modified U-Net~\citep{ronneberger2015u} where the deconvolution layers are replaced with a resizing layer followed by a convolution to mitigate the checkerboard effect. The discriminator has a fully-convolutional architecture. The two architectures are trained to minimize four loss components: $\ell_1$ reconstruction loss for the whole image, $\ell_1$ reconstruction loss for the pathological region, feature matching loss for the second to last activation layer of the discriminator, and the min-max GAN loss.}
  \label{fig:schematic}
\end{figure}

\paragraph{Dataset}

For this work, we used a dataset provided by a teledermatology service, collected in 17 clinical sites in two U.S. states from 2010 to 2018. This dataset consisted of 9897 cases and 49920 images; each case contains one or more high resolution images (resolution range: $600\times 800$ to $960\times 1280$). Ground truth of the skin condition was established for each case by an aggregated opinion of several board-certified dermatologists to differentiate among 26 common skin conditions and an additional 'other' category. It is important to note that even though the 26 skin conditions are known to be highly prevalent, the dataset itself was unbalanced, especially for certain malignant conditions like \emph{Melanoma}, which has less than 200 examples. More details on the original dataset can be found in \citep{yuanliuderm}. In addition to the skin condition, we make use of two additional pieces of information: 1) For each condition, its presence in the image is marked by a Region of Interest (ROI) bounding box (Fig.~\ref{fig:comparison}(b)) and 2) the skin color given for each case based on the Fitzpatrick skin color scale that ranges from Type I (“pale white, always burns, never tans”) to Type VI (“darkest brown, never burns”)~\citep{fitzpatrick1988validity}. Both the ROI and the skin color annotations are determined by the aggregated opinions of several dermatologist-trained annotators.

\paragraph{Data Preprocessing}

Fig.~\ref{fig:comparison}(b) shows the heterogeneous nature of this dataset. As stated previously, the region occupied by the skin condition varies significantly and the backgrounds are non-uniform and unique to each individual image (walls, hospitals, clothing, etc). As a result, the signal to noise ratio is very low in most of the images. To alleviate this problem, using the annotated ROI bounding boxes, we create a more uniform version of the dataset where the skin conditions is prominent in each image (Fig.~\ref{fig:cropping}) We devise a simple heuristic that crops a random window around an ROI or a group of adjacent ROIs while removing the presence of background information.
This results in $40000$ images of size $256\times 256$ for training the generative models and $24000$ images for evaluation.

\paragraph{Problem Formulation}

Given a set of input-output pairs $\{(x_i, m_i)\}_{i=1}^{N}$, for each image $x_i \in R^{W \times H \times C}$, $m_i \in {R^{W \times H \times {C^\prime}}}$ is its corresponding semantic map that encodes the skin color, the skin condition present in the image and the location of the condition in the image (Fig.~\ref{fig:schematic}). For a fully defined semantic map $m$, due to the possible variations (amount of hair on the skin, shooting angles, lighting conditions, morphology of the condition, etc), the corresponding image $x$ is not unique. The variations can be modeled by a conditional probability distribution $P(x|m)$. Our goal is to be able to sample from $P(x|m)$ for arbitrary and valid $m$. This image to image translation problem can be addressed using the conditional GAN framework~\citep{mirza2014conditional} which has been successfully used in 
similar settings~\citep{pix2pix, pix2pixhd, CRN, choi2018stargan}.

For each image in our dataset, the semantic map is an RGB image. The R-channel encodes the skin color and the condition is encoded in the G \& B channels by a non-zero value corresponding to its ROI bounding box(es). An example is shown in Fig.~\ref{fig:schematic}. 

Given the pairs of preprocessed skin images and their semantic maps, the problem of synthetic image generation reduces to mapping any arbitrary semantic map to a corresponding skin condition image. Pix2Pix~\citep{pix2pix} model gives a two-fold solution to this problem: An encoder-decoder architecture such as U-Net~\citep{ronneberger2015u} is trained with an $\ell_1$ reconstruction loss to reproduce a given real image from its semantic map. The main drawback, however, is that such a model produces blurry images that lack the details of a realistic image. Therefore, a second model is added to discriminate real images from synthetic ones. The addition of this min-max GAN loss results in generation of realistic images with fine-grained details. Later on, subsequent works improved the Pix2Pix method by applying various adaptations to the original algorithm: using several discriminator networks with various patch-sizes, progressively growing the size of generated images, using conditional normalization layers instead of instance normalization layers, and so forth~\citep{spade, choi2018stargan, lin2018conditional}. Similarly, in this work, based on the specifics of our data modality we apply three main adaptations to the original pix2pix algorithm:

\begin{itemize}

    \item \textbf{Checkerboard effect reduction} The original pix2pix generator implementation makes use of transposed convolution layers. As discussed by~\citet{odena2016deconvolution}, using deconvolution layers for image generation can results in ``checkerboard'' effect. The problem was resolved by replacing each deconvolution layer with a nearest-neighbor resizing layer followed by a convolution layer.
    
    \item \textbf{Condition-specific loss} The original pix2pix loss function uses the $\ell_1$ distance between the original and synthetic image as a loss function component. For skin condition images, generator model's reconstruction performance is more important in the condition ROI compared to its surrounding skin. Therefore, we add a condition-specific reconstruction term which is simply the $\ell_1$ distance between the condition ROIs in the synthetic and real images. We should mention that unlike the reconstruction loss, adding an additional condition-specific discriminator model in order to include a condition-specific min-max GAN loss did not result in improvement.
    
    \item \textbf{Feature matching loss} Feature matching loss enforces the generated images to follow the statistics of the real data through matching the features of generated and real images in a chosen layer(s) of the discriminator. It is computed as the $\ell_2$ distance between the activations of synthetic images in a chosen discriminator layer (or layers) and that of the real images. Apart from improving the quality of generated images, feature matching loss results in a more stable training trajectory~\citep{salimans2016improved}. We used the output of the discriminator's second to last convolutional layer to compute the feature matching loss.
    
\end{itemize}
    
All in all, the resulting model has four loss terms: reconstruction loss, condition-specific reconstruction loss, min-max GAN loss, and feature-matching loss. Grid-search hyperparameter selection was performed to choose the weighting coefficients for each loss component.

\section{Experiments}

\begin{figure}[h]
  \centering
  \includegraphics[width=0.8\linewidth]{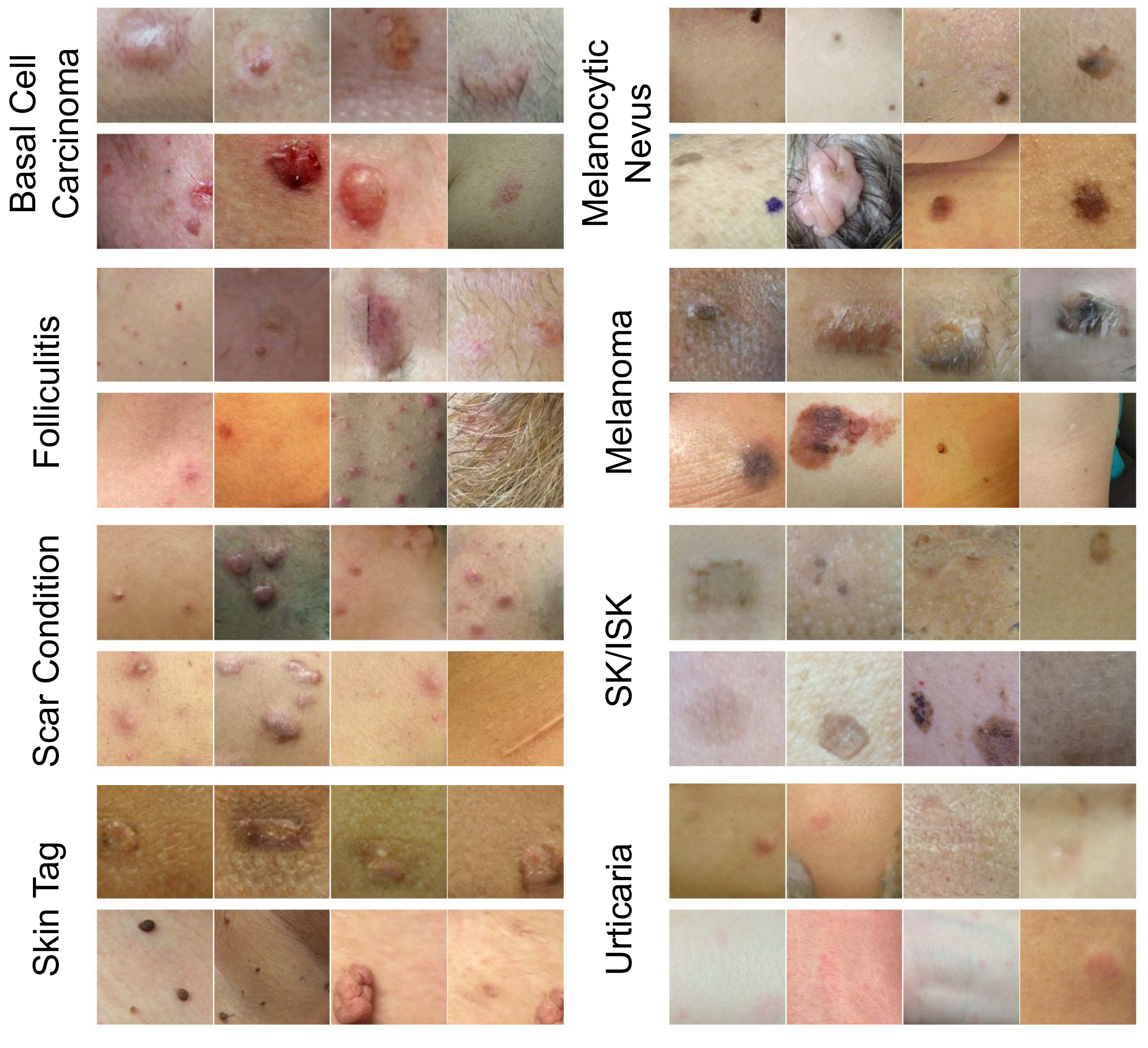}
  \caption{\textbf{Generated vs real images} For each condition, the top row shows samples of generated images and the bottom row shows samples of real images. 
  \label{fig:examples}} 
\end{figure}

Using the pre-processed dataset, we trained a DermGAN model to generate synthetic skin images with a chosen skin color, skin condition, as well as the size and region of the condition. In order to focus more on the critical and rare conditions, of the $26$ classes in the original data, we choose $8$ conditions that have fewer samples compared to other classes ($17\%$ of the dataset combined). In order to generate synthetic images, we first need to generate the corresponding semantic map. Different conditions have semantic maps with different statistics of bounding box size, shape, etc. As a result, in order to prevent domain shift between semantic maps the DermGAN is trained on and the ones it will see during test time, we use the maps in the validation set. For each synthetic image, we randomly sample a semantic map from the validation set and then apply random transformations on it: We apply random translations on the bounding boxes and if there are more than one bounding boxes in the map, we randomly select a subset of them. As a result, the same semantic map in the validation set could be used to generate diverse synthetic images. Examples of our generated images are shown in Fig.~\ref{fig:examples}.

\paragraph{Synthetic images with different skin colors}

\begin{figure}[h]
  \centering
  \includegraphics[width=0.8\linewidth]{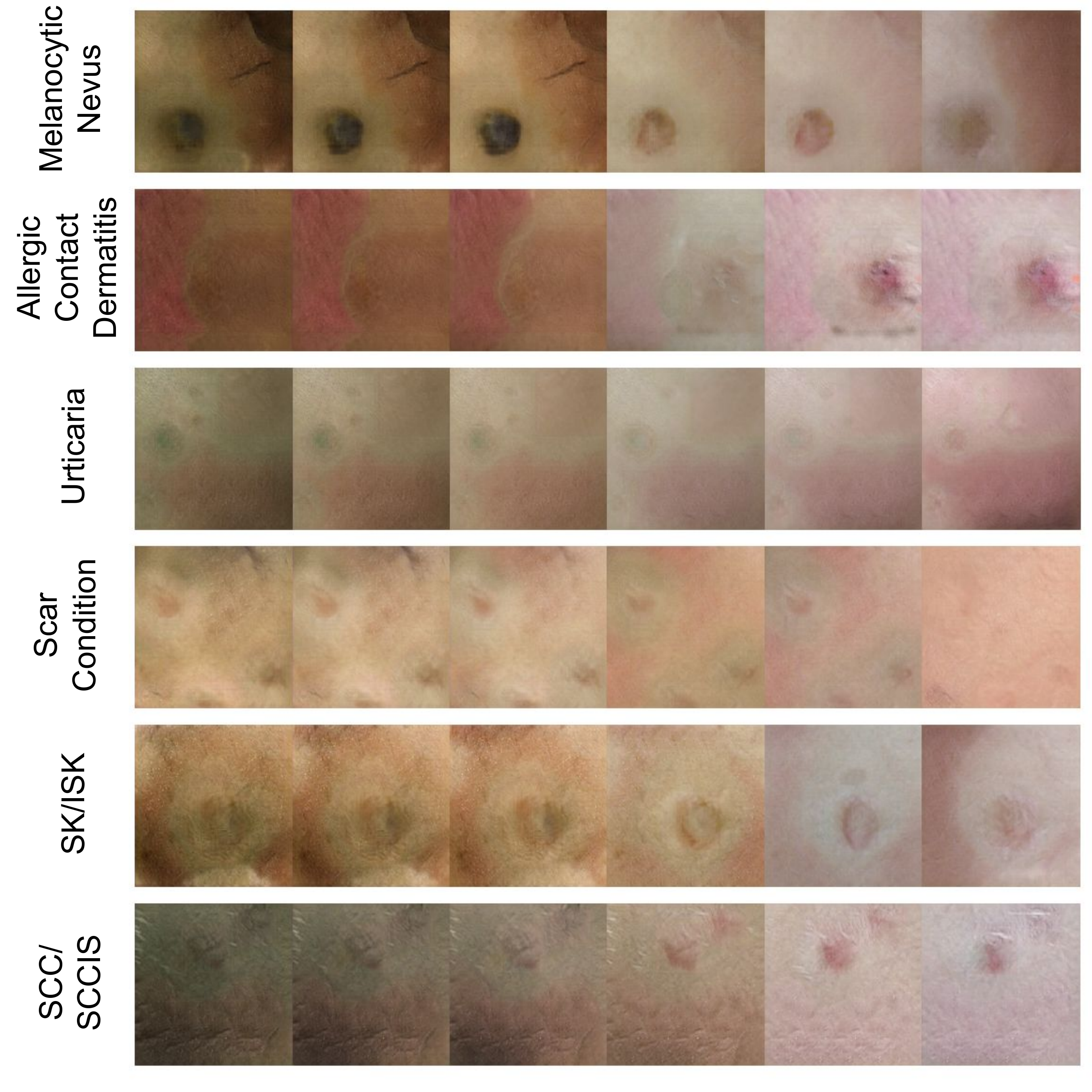}
  \caption{\textbf{Skin color variation} By changing the encoded skin color in the semantic map, we are able to create synthetic images with diverse skin colors. It should be noted that as the skin color varies, the change in surrounding visuals is consistent with real world occurrences.
  \label{fig:tone}} 
\end{figure}

In this and the subsequent experiment, we train a DermGAN model on all of the $26$ conditions of the dataset to use images of wider demographics. For a given semantic map in the test set, we vary the encoded background color and observe the respective changes in the generated image. Fig.~\ref{fig:tone} depicts examples of this experiment, in which the encoded skin color of a semantic map is replaced with each of the six types. As illustrated in the figure, the DermGAN model is able to change the background skin color while adjusting the condition itself to reflect this change. For instance, for \emph{Melanocytic Nevus}, the generated image for the darker tones has also a darker mole, which mimics real data.

\paragraph{Synthetic images with different sizes of the skin condition}

\begin{figure}[h]
  \centering
  \includegraphics[width=0.8\linewidth]{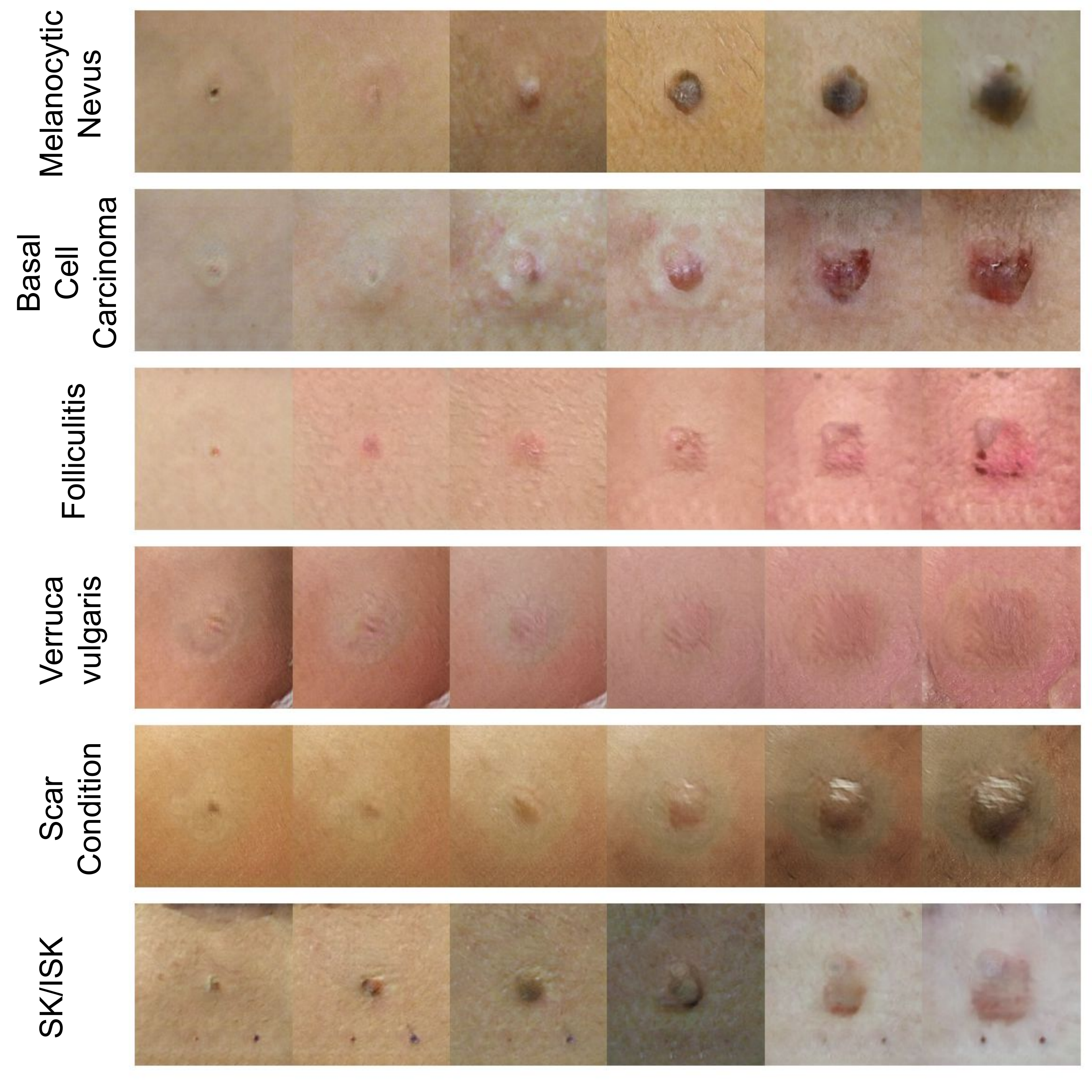}
  \caption{\textbf{Skin condition size variation} As DermGAN generates synthetic images given the condition's ROI, we are able to generate the same condition with various sizes. For specific conditions (e.g. \emph{Basal Cell Carcinoma}), the change in size results in a change in the surrounding visuals, which is consistent with the real world occurrences.
  \label{fig:size}} 
\end{figure}

 For a given semantic map, we can vary the sizes of the pathological region for each skin condition and observe the respective changes in the generated image. Fig.~\ref{fig:size} shows examples of this experiment, in which the size of the bounding box of a semantic map is gradually increased. We observe that as the size of the skin condition changes, the visual appearance also changes, which is consistent with real world occurrences. Note that in this experiment, the semantic maps we fed into the model are generated synthetically (not by applying random transformations on semantic maps in the validation dataset.)

\paragraph{GAN evaluation metrics}
A perfect objective evaluation of GAN generated images remains a challenge~\citep{theis2015note}. One widely-used measure is the inception score~\citep{salimans2016improved} that works as a surrogate measure of the diversity and the amount of distinct information in the synthetic images. It is computed as the average KL-divergence between the class probabilities assigned to a synthetic sample by an Inception-V3 model~\citep{russakovsky2015imagenet} trained on the ImageNet dataset~\citep{szegedy2016rethinking} and the average class probabilities of all synthetic samples. The main drawback that makes the use of inception score inadmissible in our case is that it assumes the classes in the dataset at hand to be a subset of the 1000 ImageNet classes~\citep{barratt2018note}. Another widely-used measure is the Frechet Inception Distance (FID)~\citep{heusel2017gans}. FID directly measures the difference between the distribution of generated and real images in the activation space of the ``Pool 3'' layer of the Inception-V3 model. We perform an ablation study of the DermGAN model. Results of the FID scores on our test set (24000 images) are reflected in Table~\ref{table:ablation} (confidence intervals are for 50 trials).

\newcommand{\specialcell}[2][c]{\begin{tabular}[#1]{@{}c@{}}#2\end{tabular}}
\begin{table}[b]
\centering
    \small
    \begin{tabular}{c|c|c|c|c|c}
          & \specialcell{Real\\Data}  & \specialcell{Derm\\GAN}  & \specialcell{No\\checkerboard\\effect\\ mitigation} & \specialcell{No\\ condition-specific\\reconstruction\\ loss} & \specialcell{No\\ feature\\matching\\loss} \\ 
        \hline
        \vspace{3mm}
        FID ($\pm$ 1.96 STD)              & 83.6 $\pm$ 2.5 & 122.4 $\pm$ 3.4 & 151.6 $\pm$ 3.4 & 174.0 $\pm$ 4.7 & 140.7 $\pm$ 2.5 \\

        
    \end{tabular}
    \vspace{2mm}
     \caption{\textbf{Ablation study of GAN evaluation using FID scores.}\label{table:ablation} Note that a lower FID score means a better performance.}
\end{table}

\paragraph{Human Turing test}

\begin{figure}[h]
  \centering
  \includegraphics[width=\linewidth]{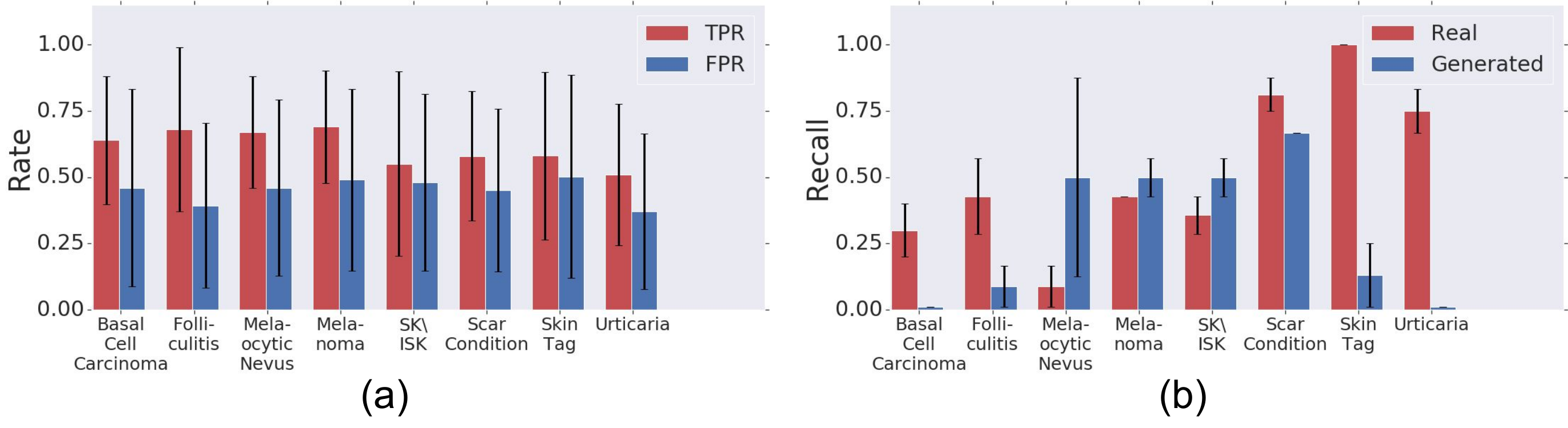}
  \caption{\textbf{Human Turing test} Results for discriminating between real and synthetic images are shown in (a), whereas results for whether images correctly describe the respective skin condition are shown in (b). Error bars represent standard deviation.
  \label{fig:turing}} 
\end{figure}

For a subjective measure of how realistic the generated images are, we conducted two qualitative experiments. The first test was a Turing test with $10$ participants. Each participant was asked to choose the skin images they found realistic in a collection of 80 real and 80 randomly selected synthetic images. On average the true positive rate (TPR) (the ratio of real images correctly selected) is $0.52$  and the false positive rate (FPR) (the ratio of synthetic images detected as real) is $0.30$. Results for each condition are demonstrated in Fig.~\ref{fig:turing}(a), with average TPR ranging from $0.51$ to $0.69$ and average FPR from $0.37$ to $0.50$. As expected, the TPR is higher than FPR for all conditions. However, the high FPR rate among all conditions indicate the high fidelity of synthetic images. The standard deviation of participants' performances is large  and we hypothesize that it is due to the fact that they come from various backgrounds with different levels of experience with skin images.
 
The second experiment was designed to measure the medical relevance of the synthetic images. In this experiment, two board certified dermatologists answered a set of 16 questions. In each question, the participants were asked to choose the images relevant to a given skin condition among a combined set of real and randomly selected synthetic images. The average recall (ratio of related images correctly chosen) is $0.61$ and $0.45$ for the real and synthetic images respectively. Results for each condition are shown in Fig.~\ref{fig:turing}(b), with recall ranging from $0.3$ to $1.00$ for real images and from $0.00$ to $0.67$ for synthetic images. For \emph{Melanocytic nevus}, \emph{Melanoma}, and \emph{Seborrheic Keratosis / Irritated Seborrheic Keratosis (SK/ISK)}, synthetic images were identified to better represent the respective skin condition, indicating that our approach is able to preserve the clinical characteristics of those skin conditions.

\begin{figure}[h]
  \centering
  \includegraphics[width=0.8\linewidth]{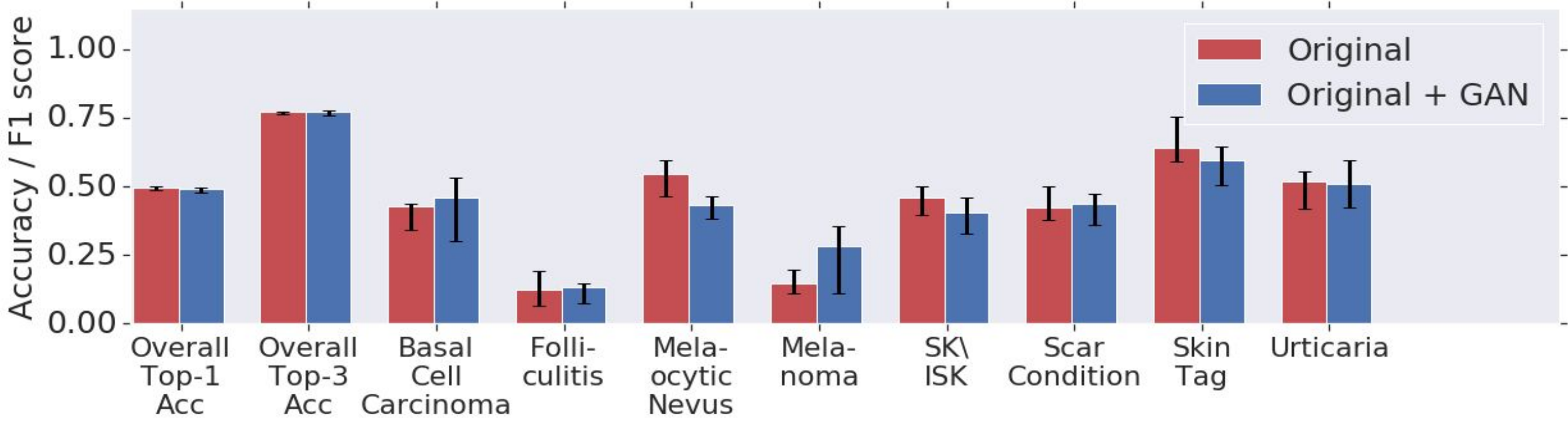}
  \caption{\textbf{Augmenting training data with synthetic images.}
  We added $20000$ synthetic images to the original training data of size $49920$. The overall performance is comparable to the baseline, but the performance on rare conditions like \emph{Melanoma} and \emph{Basal cell carcinoma} has noticeable improvement.
  \label{fig:addition}} 
\end{figure}

\paragraph{Synthetic images as data augmentation for skin condition classification}
We first trained a MobileNet model~\citep{howard2017mobilenets} on our original (uncropped) data to differentiate between $27$ skin condition classes ($26$ plus ``other'') from a single image. This baseline model achieves a top-1 accuracy of $0.496$ on a test set of 5206 images, with poor performance on some of the rare conditions. To help alleviate this issue, we generate 20000 synthetic images using the 8-class DermGAN model and add them to the existing training data. We trained another MobileNet skin condition classifier using this enriched dataset and evaluated its performance on the same test set. While the top-1 accuracy remains relatively unchanged (p = $0.56$ using paired T-test), performance improves for some of the malignant minority classes: \emph{Melanoma} $F1$ score increases from $0.148$ ([$0.067$, $0.193$], $95\%$ confidence interval using bootstrapping) to $0.282$ ([$0.110$, $0.356$]), whereas \emph{Basal cell carcinoma} $F1$ score increases from $0.428$ ([$0.343$, $0.439$]) to $0.458$ ([$0.301$, $0.534$]), though at the cost of misclassifying \emph{Melanocytic nevus} (0.113 decrease in $F1$). For the other $5$ classes, the performances between the two models are comparable. (Fig.~\ref{fig:addition}). Conventional data augmentation techniques (flipping, saturation, jitters) are used in both of the training setups.

\section{Discussion and Conclusion}
In this work, we tackle the problem of generating clinical images with skin conditions as seen in a teledermatology setting. We frame the problem as an image to image translation task and propose DermGAN, an adaptation of the popular Pix2Pix GAN architecture. Using the proposed framework we are able to generate realistic images for pre-specified skin condition. We demonstrate that when varying the skin color or the size and location of the condition, the synthetic images can reflect such changes, while maintaining the characteristics of the respective skin condition. We further demonstrate that our generated images are of high fidelity using objective GAN evaluation metrics and qualitative tests. When using the synthetic images as data augmentation for training a skin condition classifier, the model is comparable to baseline, with improved performance on rare skin conditions. Further work is needed to improve the resolution and the diversity of the generated images, to effectively utilize such images for classification tasks by using techniques such as~\citep{grover2019bias}, and to explore the benefit of image synthesis in other clinical applications.

\acks{The authors would like to acknowledge Susan Huang and Kimberly Kanada for their clinical support. Thanks also go to Yun Liu, Greg Corrado, and Erica Brand for their feedback on the project. We also appreciate the help from the rest of the dermatology research team at Google Health for their engineering and operational support.}

\clearpage{}

\bibliography{refs}

\bibliographystyle{aaai}

\end{document}